\documentclass[a4paper]{article}

\usepackage{iclr2019_conference,times}

\usepackage[T1]{fontenc}    % use 8-bit T1 fonts
\usepackage{url}            % simple URL typesetting
\usepackage{booktabs}       % professional-quality tables
\usepackage{amsmath}
\usepackage{amsfonts}       % blackboard math symbols
\usepackage{nicefrac}       % compact symbols for 1/2, etc.
\usepackage{microtype}      % microtypography
\usepackage{graphicx}
\usepackage[colorlinks=true, allcolors=blue]{hyperref}
\usepackage[english]{babel}
\usepackage[utf8x]{inputenc}
\usepackage{EvolvingAIcommenting}
\newif\ifcomments

\commentsfalse
\ifcomments
\newcommand{\comments}[1]{#1}
\else
\newcommand{\comments}[1]{}
\fi
\iclrfinalcopy

\author{
\centerline{Thomas Miconi\thanks{Correspondence: \texttt{tmiconi@uber.com}}, Aditya Rawal, Jeff Clune \& Kenneth O. Stanley}\\
\centerline{Uber AI Labs}\\
\centerline{\texttt{tmiconi|aditya.rawal|jeffclune|kstanley@uber.com}}
}

\title{Backpropamine: training self-modifying neural networks with differentiable neuromodulated plasticity}
\date{}

\begin{document}

\maketitle

\begin{abstract}

%\ks{maybe have an intro sentence on why we want plasticity (or even what it is) in the first place?}

The impressive lifelong learning in animal brains is primarily enabled by plastic changes in synaptic connectivity. Importantly, these changes are not passive, but are actively controlled by neuromodulation, which is itself under the control of the brain. The resulting self-modifying abilities of the brain play an important role in learning and adaptation, and are a major basis for biological reinforcement learning. Here we show for the first time that artificial neural networks with such neuromodulated plasticity can be trained with gradient descent. Extending previous work on differentiable Hebbian plasticity, we propose a differentiable formulation for the neuromodulation of plasticity. 
%We implement two forms of neuromodulation: a simple form that merely puts plasticity rates under network control, and a more complex one that emulates the retroactive properties of dopamine on synaptic plasticity, effectively leveraging Hebbian plasticity and neuromodulation to implement eligibility traces.https://v2.overleaf.com/18311816qbwyhqwfwsqr
We show that neuromodulated plasticity improves the performance of neural networks on both reinforcement learning and supervised learning tasks. In one task, neuromodulated plastic LSTMs with millions of parameters outperform standard LSTMs on a benchmark language modeling task (controlling for the number of parameters). We conclude that differentiable neuromodulation of plasticity offers a powerful new framework for training neural networks.
\jc{this scopes the impact to only a new way to make *self-modifying* networks, so someone has to care about that (most don't). Should we instead say ``to train powerful neural networks''?}\ks{I reworded it to address Jeff's concern.}
%\ks{note if Aditya's result gets in the last sentence should be modified}
%The ability to allow neural networks to self-determine the level of plasticity of different components over their lifetime opens interesting avenues for future research. 
%\ks{can we be more specific about what the "interesting avenues" actually are?  this concluding sentence is suspiciously vague here.}

%Training neural networks to self-modify, the way Nature intended.
\end{abstract}

\section{Introduction}

%\ks{add command to sort multi-citations in numerical order}

Neural networks that deal with temporally extended tasks must be able to store traces of past events.
%, whether as item memories or procedural adaptations
\jc{I have no idea what this last clause means}\ks{good point, also found this sentence difficult to parse}\tm{Deleted.}
Often this memory of past events is maintained by neural activity reverberating through recurrent connections; other methods for handling temporal information exist, including memory networks \citep{sukhbaatar15end} or temporal convolutions \citep{mishra2017simple}. However, in nature, the primary basis for long-term learning and memory in the brain is \emph{synaptic plasticity} -- the automatic modification of synaptic weights as a function of ongoing activity \citep{martin2000synaptic,liu2012optogenetic}. Plasticity is what enables the brain to store information over the long-term about its environment that would be impossible or impractical for evolution to imprint directly into innate connectivity (e.g. things that are different within each life, such as the language one speaks).

Importantly, these modifications are not a passive process, but are
actively modulated on a moment-to-moment basis by dedicated systems and
mechanisms: the brain can ``decide'' where and when to modify its own
connectivity, as a function of its inputs and computations. This \emph{neuromodulation} of plasticity, which involves several chemicals (particularly dopamine; \citealt{calabresi2007dopamine,he2015distinct,li2003dopamine,yagishita2014critical}), plays an important role in learning and adaptation \citep{molina-luna2009dopamine,smith-roe2000-us,kreitzer2008striatal}. By allowing the brain to control its own modification as a function of ongoing states and events, the neuromodulation of plasticity can filter out irrelevant events while selectively incorporating important information, combat catastrophic forgetting of previously acquired knowledge, and implement a self-contained reinforcement learning algorithm by altering its own connectivity in a reward-dependent manner \citep{schultz1997neural,niv2009reinforcement,frank2004carrot,hoerzer2014emergence,miconi2016biologically,ellefsen2015neural,velez2017diffusion}.

The complex organization of neuromodulated plasticity is not accidental: it results from a long process of evolutionary optimization. Evolution has not only designed the general connection pattern of the brain, but has also sculpted the machinery that controls neuromodulation, endowing the brain with carefully tuned self-modifying abilities and enabling efficient lifelong learning. In effect, this coupling of evolution and plasticity is a meta-learning process (the original and by far most powerful example of meta-learning), \ks{switched "most successful" to "most iconic" here.  I'm not persuaded by Jeff's argument (now commented out above) that  "most agree" with it being successful -- my issue is not with people disagreeing, but rather with it being understated.  calling human-level learning the "most successful" greatly understates how amazing it is, as if it merits even comparing to something like A3C.}\tm{I'm really not sure about 'iconic' here, but I'll see what Jeff says}\jc{I am happy to add even more hyperbole, but iconic doesn't seem to do that. I still prefer most successful to iconic (somehow iconic to me seems to give even less weight and seems out of place), but am open to some other word or phrase to highlight how much more successful we mean than current meta-learning}\ks{tried a new rephrasing. still don't like "most successful" as it makes it sound like there's even a comparison.}
\jc{Now it makes it sound even more scoped. It is the best example in Nature, not the best example overall (comparing Nature to ML). It's also confusing: are there other examples in nature? How about this ``(the original and by far most powerful example of meta-learning)''}
\tm{Went with Jeff's version}whereby a simple but powerful optimization process  (evolution guided by natural selection) discovered how to arrange elementary building blocks to produce remarkably efficient learning agents.

%The importance of neuromodulation for biological learning is particularly apparent in the context of reinforcement learning based on rewards and punishments. %Neuromodulation of plasticity is thought to be one of the main bases of reinforcement learning \citep{NIV, frank2004carrot,fiete2007model}; conversely, the %neuromodulatory systems of the brain (especially the dopaminergic system) are important source of inspiration for modern reinforcement learning algorithm, %including temporal difference learning and actor-critic algorithms such as A3C (which make use of value prediction, reward prediction error and use these to %modulate weight changes). The REINFORCE algorithm itself can be implemented in reward-modulated Hebbian plasticity (see e.g. Eq. 11 in %\citep{williams92reinforce}) (However, there are important caveats in implementing RL with reward-modulated Hebbian learning, discussed by several papers from Fremaux and Gerstner). Therefore, endowing artificial neural networks with self-controlled neuromodulated plasticity might allow them to develop their own %self-contained reinforcement learning algorithm, acting not just on the (necessarily volatile) hidden recurrent state, but directly on the connectivity on the %network.

Taking inspiration from nature, several authors have shown that evolutionary algorithms can design small neural networks (on the order of hundreds of connections)
with \jc{why have self-controlled here? It potentially makes the reader thing that there is a subset of NM called self-controlled NM. I think it's clearer just to define neuromodulated plasticity above (including that it is self-modifying) and then use just the words ``neuromodulated plasticity'' from then on out}\ks{agreed}\tm{Deleted} neuromodulated plasticity (see the ``Related Work'' section below). However, many of the spectacular recent advances in machine learning make use of gradient-based methods (which can directly translate error signals into weight gradients) rather than evolution (which has to discover the gradients through random weight-space exploration). If we could make plastic, neuromodulated networks amenable to gradient descent, we could leverage gradient-based methods for optimizing and studying neuromodulated plastic networks, expanding the abilities of current deep learning architectures to include these important biologically inspired self-modifying abilities.

Here we build on the differentiable plasticity framework \citep{miconi2016biologically,miconi2018differentiable} to implement differentiable neuromodulated plasticity. As a result, for the first time to our knowledge, we are able to train neuromodulated plastic networks with gradient descent. We call our framework \emph{backpropamine} in reference to its ability to emulate the effects of natural neuromodulators (like dopamine) in artificial neural networks trained by backpropagation. 
%Two forms of neuromodulation are implemented, demonstrating the flexibility of the approach: a simple form that directly
%modulates plasticity rates, and a slightly more complex form inspired by the retroactive action of dopamine on plasticity, which effectively implements eligibility traces, an important tool in reinforcement learning \citep{sutton1998reinforcement}.
\jc{(an important tool in RL [cite])}\tm{Cutting it out to save space} 
Our experimental results establish that
neuromodulated plastic networks outperform both non-plastic and non-modulated plastic networks,
both on simple reinforcement learning tasks and on a complex language modeling task involving a multi-million parameter network. By showing that neuromodulated plasticity can be optimized through gradient descent, the backpropamine framework potentially provides more powerful types of neural networks, both recurrent and feedforward, for use in all the myriad domains in which neural networks have had tremendous impact.
%opens exciting avenues for future research on self-modifying networks and potentially provides more powerful types of neural networks, both recurrent and feedforward, for use in all the myriad domains in which neural networks have had tremendous impact. 
\jc{and potentially provides more powerful types of neural networks, both recurrent and feedforward, for use in all the myriad domains in which neural networks have had tremendous impact}\ks{addition sounds good to me}\tm{Done}
%\ks{reorganized several sentences in this paragraph because the subject of every single sentence in the whole paragraph was "We."  Even if we like writing biographically, it's not a desirable style to start every sentence with the same subject for 6 sentences in a row.} 

%\ks{again will need to think how to highlight Aditya's result as well in the intro.  it's a very nice result if it holds up and something we'd want to advertise in addition to "simple reinforcement learning tasks"}

\section{Related work}

Neuromodulated plasticity has long been studied in evolutionary computation. Evolved networks with neuromodulated plasticity were shown to outperform both non-neuromodulated and non-plastic networks in various tasks (e.g. \citealt{soltoggio2008evolutionary,risi2012unified}; see \citealt{soltoggio2017born} for a review). A key focus of neuromodulation in evolved networks is the mitigation of \emph{catastrophic forgetting}, that is, allowing neural networks to learn new skills without overwriting previously learned skills. By activating plasticity only in the subset of neural weights relevant for the task currently being performed, knowledge stored in other weights about different tasks is left untouched, alleviating catastrophic forgetting \citep{ellefsen2015neural,velez2017diffusion}. However, evolved networks were historically relatively small and operated on low-dimensional problem spaces.  

%In previous work, we showed how the plasticity of individual synaptic connections can
%be optimized by gradient descent, in the same way that standard synaptic weights
%are \citep{miconi2016backpropagation,miconi2018differentiable}. However, 
The differentiable plasticity framework \citep{miconi2016backpropagation,miconi2018differentiable} allows the plasticity of individual synaptic connections to
be optimized by gradient descent, in the same way that standard synaptic weights
are. However, 
%\ks{added this following small bit of text before the next comma to say something positive for the method} 
% OK
while it could improve performance in some tasks over recurrence without plasticity, this method only facilitated passive, non-modulated plasticity, in which weight changes occur automatically as a function of pre- and post-synaptic activity. Here we
extend this framework to implement differentiable neuromodulated plasticity, in which the
plasticity of connections can be modulated moment-to-moment through a signal computed by the network. This extension allows the network itself to decide over its lifetime 
where and when to be plastic, endowing the network with true self-modifying abilities.

There are other conceivable though more complex approaches for training self-modifying networks. For example, the weight modifications can themselves be computed by a neural network \citep{schmidhuber1993self,schlag2017gated,munkhdalai2017meta,wu2018meta}. 
%\ks{Maybe the following sentence could be reworded to avoid creating an expectation (we even use the word "expect") of a comparison.  The way it is worded now (which seems further emphasized by being the final sentence of the section), it creates an expectation that there is a direct comparison forthcoming with Schmidhuber's style of network-modifying-a-network.} 
% Done - put 'suggest' instead
However, none so far have taken the simple approach of directly optimizing the neuromodulation of plasticity itself within a single network, through gradient descent instead of evolution, as investigated here.

\ks{Notice that Jeff (below) had exactly the same worry I did about the sentence that previously concluded the paragraph above, which makes me even more worried (since it confirms my worry).  While Jeff below suggests simply conceding we don't investigate the hypothesis in this paper, I think we should simply not advance any hypothesis that leaves the reader expecting experiments we do not provide.  Therefore, I concluded the above paragraph with entirely new text that does not imply any necessary empirical comparison and I cut the old concluding text enitrely (though I left it in a comment in the source below)}%We suggest that leveraging the powerful mechanism of neuromodulated Hebbian %plasticity may result in more compact (in terms of trainable parameters) yet %powerful architectures for self-modifying networks
\jc{, although we do not investigate that hypothesis in this work.}\jc{I added this comment because we do not want to set expectations that we investigate that and then fail to meet those expectations}\jc{EDIT: I think Ken's current solution is safer.}\tm{Great!}

\section{Methods}

\subsection{Background: Differentiable Hebbian plasticity} 
 
The present work builds upon the existing differentiable plasticity framework \citep{miconi2016backpropagation,miconi2018differentiable}, which allows gradient descent to optimize not just the weights, but also the plasticity of each connection. In this framework, each connection in the network is augmented with a Hebbian plastic
component that grows and decays automatically as a result of ongoing activity. In effect, each connection contains a \textcolor{green}{fixed} and a \textcolor{red}{plastic} component:

%\begin{multline}
%    x_j(t) = \sigma \big\{\sum_{i \in inputs}  [ w_{i,j}  x_i(t-1) \\ + \alpha_{i,j} \mathrm{Hebb}_{i,j}(t)  x_i(t-1) ] \big\},
%\end{multline}
\begin{equation}
    x_j(t) = \sigma \big\{\sum_{i \in \text{inputs to $j$}}  (\textcolor{green}{w_{i,j}}  + \textcolor{red}{\alpha_{i,j} \mathrm{Hebb}_{i,j}(t)})  x_i(t-1)  \big\} \label{eq:network}
\end{equation}
\begin{equation}
    \mathrm{Hebb}_{i,j}(t+1) = \mathrm{Clip}(\mathrm{Hebb}_{i,j}(t) + \eta  x_i(t-1)x_j(t)),  \label{eq:hebb}        
\end{equation}
%\ks{I believe Clip should also be mathrm style -- it's a function just as Hebb is right?}
% Done!

where $x_i(t)$ is the output of neuron $i$ at time $t$, $\sigma$ is a nonlinearity (we use $\tanh$ in all experiments), $w_{i,j}$ is the baseline (non-plastic) weight of the connection between neurons $i$ and $j$, and $\alpha_{i,j}$ is the  \emph{plasticity coefficient} that scales the magnitude of the plastic component of the connection. The plastic content is represented by the \emph{Hebbian trace} $\mathrm{Hebb}_{i,j}$, which accumulates the product of pre- and post-synaptic activity at connection $i,j$, as shown in Eq. \ref{eq:hebb}. 

$\mathrm{Hebb}_{i,j}$ is initialized to zero at the beginning of each episode/lifetime, and is updated automatically according to Eq. \ref{eq:hebb}: it is a purely episodic/intra-life quantity. By contrast, $w_{i,j}$, $\alpha_{i,j}$ and $\eta$ are the structural components of the network, which are optimized by gradient descent between episodes/lifetimes to minimize the expected loss over an episode. 

The function $\mathrm{Clip}(x)$ %\ks{use mathrm here too for Clip} 
in Eq. \ref{eq:hebb} is any function or procedure that constrains $\mathrm{Hebb}_{i,j}$ to the $[-1, 1]$ range, to negate the inherent instability of Hebbian learning. In previous work \citep{miconi2018differentiable}, this function was either a simple decay term, or a normalization implementing Oja's rule \citep{oja2008oja}. In the present paper it is simply a hard clip ($x \leftarrow 1$ if $x>1$; $x \leftarrow -1$ if $x<-1$).
%, or a soft clip ($x \leftarrow x + \Delta x * (1-x)$ if $\Delta x > 0$; $x \leftarrow x + \Delta x * (1 + x)$ if $\Delta x < 0$). 
Compared to previously used operations, this simple operation turned out to produce equal or superior performance on the tasks in this paper.

Note the distinction between the $\eta$ and $\alpha_{i,j}$ parameters: $\eta$ is the intra-life ``learning rate'' of plastic connections, which determines how fast new information is incorporated into the plastic component, while $\alpha_{i,j}$ is a scale parameter, which determines the maximum magnitude of the plastic component (since $\mathrm{Hebb}_{i,j}$ is constrained to the [-1,1] range).
%While $\eta$ could also be independently optimized for each connection, like the $\alpha_{i,j}$, we use a single, global $\eta$ for simplicity.

\jc{Importantly, in contrast to other approaches using uniform plasticity \citep{schmidhuber1993reducing} or the ``or'' here is confusing. are these two different things, or are you saying that uniform plasticity is also known as fast weights? if the latter, remove the or and put ``uniform plasticity (cite), also known as ``fast weights'' (and get rid of the so-called)}  \jc{to save time, I just went ahead and made the suggestion below.}
Importantly, in contrast to other approaches using \emph{uniform} plasticity \citep{schmidhuber1993reducing}, including \jc{should this instead be ``which includes'') i.e. is fast weights exactly the same as uniform weights, or it is an example of it (i.e. a subset of it)?} ``fast weights'' \citep{ba2016fast}, the amount of plasticity in each connection (represented by $\alpha_{i,j}$) is \emph{trainable}, allowing the meta-optimizer to design complex learning strategies (see \citealt{miconi2018differentiable} for a discussion of this point, and experimental comparisons that demonstrate and explain superior performance of differentiable plasticity over uniform-plastic networks).\tm{Done. Note: Fast weights are a (complicated) subset of uniform-plastic network. I don't exactly test against the actual, full-fledged fast-weight algorithm, so I removed mention of it in this parenthesis.}

%\ks{Should we mention something about how easy plasticity is to program/include in modern DL platforms?  One might come to conclusion from the section above that plasticity is theoretically interesting but practically so different from contemporary DL that it would just be a big inefficient morasse to implement.  But it's actually very simple (as we noted in the ICML paper), and we might also want to mention that here as well.}\jc{I like that idea}

An important aspect of differentiable plasticity is extreme ease of implementation: implementing a plastic recurrent network only requires less than four additional lines of code over a standard recurrent network implementation \citep{miconi2018differentiable}. The Backpropamine framework described below inherits this simplicity; in particular, the ``simple neuromodulation'' approach does not require any additional code over differentiable plasticity, but merely a modification of it.

%\ks{I think I said something similar previously (though it may have been cut due to our space limitations), but I think we need to say something here about how differentiable plasticity added value in prior work over and above recurrence alone.  Otherwise, it doesn't make sense why anyone should care when we (soon enough below) compare differentiable plasticity without neuromodulation to with neuromodulation.  That is, who cares whether we beat a method that may be useless to begin with?  We need to let the reader know that differentiable plasticity actually does something useful to justify comparing only to it in most experiments (other than Aditya's) coming up.}
%\tm{We now include results with non-plastic networks}

\subsection{Backpropamine: Differentiable neuromodulation of plasticity}

Two methods are proposed to introduce neuromodulated plasticity within the differentiable plasticity framework. In both cases, plasticity is modulated on a moment-to-moment basis by a network-controlled neuromodulatory signal $M(t)$. The computation of $M(t)$ could be done in various ways; at present, it is simply a single scalar output of the network, which is used either directly (for the simple RL tasks) or passed through a meta-learned vector of weights (one for each connection, for the language modeling task). 
We now explain  how the equations of differentiable plasticity are modified to make use of this neuromodulatory signal.

\subsubsection{Simple neuromodulation}
The simplest way to introduce neuromodulation of plasticity in this framework is to make the (global) $\eta$ parameter depend on the output of one or more neurons in the network. Because $\eta$ essentially determines the rate of plastic change, placing it under network control allows the network to determine how plastic connections should be at any given time. Thus, the only modification to the equations above in this \emph{simple neuromodulation} variant is to replace $\eta$ in Eq. \ref{eq:hebb} with the network-computed, time-varying neuromodulatory signal $M(t)$. That is, Eq. \ref{eq:hebb} is replaced with

\begin{equation}
    \mathrm{Hebb}_{i,j}(t+1) = Clip(\mathrm{Hebb}_{i,j}(t) + M(t)  x_i(t-1)x_j(t)). \label{eq:simplemodulation}
\end{equation}

\subsubsection{Retroactive neuromodulation and eligibility traces}

\ks{This section is definitely a surprising amount of text and effort if it never helps in any experiments.  Very good if Aditya's experiment ends up showing a benefit for such traces.}
\tm{True, but it shows the flexibility of the approach, and there may be domains where eligibility traces have more impact. (Also, it's cool!)}\jc{Ken, see my comment on this in the printed-version edits I sent out}\ks{agreed with both -- we should still keep the section, but also agree with Jeff's printed comment that it would help it make a little more sense if we briefly say something in discussion about how this kind of trace is likely important for other purposes (see Jeff's comment).}

More complex schemes are possible. In particular, we introduce an alternative neuromodulation scheme that takes inspiration from the short-term retroactive effects of neuromodulatory dopamine on Hebbian plasticity in animal brains. In several experiments, dopamine was shown to retroactively gate the plasticity induced by \emph{past} activity, within a short time window of about 1s \citep{yagishita2014critical,he2015distinct,fisher2017reinforcement,cassenaer2012conditional}. Thus, Hebbian plasticity does not directly modify the synaptic weights, but creates a fast-decaying ``potential'' weight change, which is only incorporated into the actual weights if the synapse receives dopamine within a short time window. As a result, biological Hebbian traces essentially implement a so-called \emph{eligibility trace} \citep{sutton1998reinforcement}, keeping memory of which synapses contributed to recent activity, 
%while the dopamine signal produces a reward prediction error signal that modulates actual plastic changes, such that weight changes only affect recently active connections when an unpredicted reward occurs.
\ks{in what sense is it "unexpected"?}\tm{Replaced "unexpected" with "unpredicted" to keep with the previous sentence}\ks{Okay, but what I meant was that it may be a bit confusing how the dopamine signal represents a "reward prediction error" -- where is the "predicted" reward in this system and how is the "error" computed off it (intuitively)?}\jc{I think the confusion arises because it is not entirely clear if this sentence is describing what happens in natural brains or in our model. If the former, we know it is about unpredicted reward. But if it is our model, then the system could learn to do that (which is cool to point out), but of course we don't know if it does do that. So perhaps we could reframe it that way: in nature, what happens is x. our system is cool because it could learn to do the same, but of course it may learn to do something else too. Studying to what extent our artificial NM systems converge to the same solution found in animal brains is an interesting path for future research. Would that address your issues Ken?. }\ks{that is an improvement}\tm{Adding a few words to clarify that this entire paragraph is about what happens in the brain.}\jc{That makes the above clearer, but the first sentence of the next paragraph just raises all Ken's questions again, because the reader may not know how our model involves making predictions, checking if they were wrong, and then turning on dopamine if they were. See my suggestion in the email thread. We may want to say that our system could do this because M(t) is a learned function of the past and present, but of course it could learn to do something else.} 
\tm{This is a model of the dopamine effects on plasticity, not how dopamine itself is delivered. I changed the wording and commented out some stuff to concentrate on that}
while the dopamine signal modulates the transformation of these eligibility traces into actual plastic changes.
Such mechanisms have been modelled in computational neuroscience studies, e.g. \citep{izhikevich2007solving,hoerzer2014emergence,fiete2007model,soltoggio2013solving,miconi2016biologically} (see \citealt{gerstner2018eligibility} for a recent review of this concept).  

Our framework easily accommodates this more refined model of dopamine effects on plasticity. We simply replace Eq.\ \ref{eq:hebb} above with the two equations,

\begin{equation}
    \mathrm{Hebb}_{i,j}(t+1) = Clip(\mathrm{Hebb}_{i,j}(t) + M(t) E_{i,j}(t) )  \label{eq:retroactive1}
\end{equation}
\begin{equation}
    E_{i,j}(t+1) = (1-\eta)E_{i,j}(t) + \eta  x_i(t-1)x_j(t)). \label{eq:retroactive2}  
\end{equation}

Here $E_{i,j}(t)$ (the eligibility trace at connection $i,j$) is a simple exponential average of the Hebbian product of pre- and post-synaptic activity, with trainable decay factor $\eta$. $\mathrm{Hebb}_{i,j}(t)$, the actual plastic component of the connection (see Eq. \ref{eq:network}), simply accumulates this trace, but gated by the current value of the dopamine signal $M(t)$. Note that $M(t)$ can be positive or negative, approximating the effects of both rises and dips in the baseline dopamine levels \citep{schultz1997neural}. 

\section{Experiments}

\subsection{Task 1: Cue-reward association}

\begin{figure}
\begin{center}
\hspace*{-1cm}
\includegraphics[scale=.37]{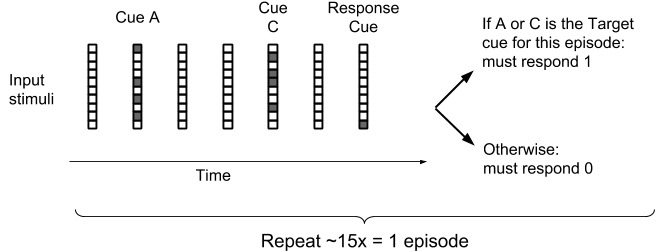} % https://docs.google.com/drawings/d/11zSnnVLhr4mydBZTjMRDCij9rPGhTy6DL2PdGM3_WUk/edit
\includegraphics[scale=.38]{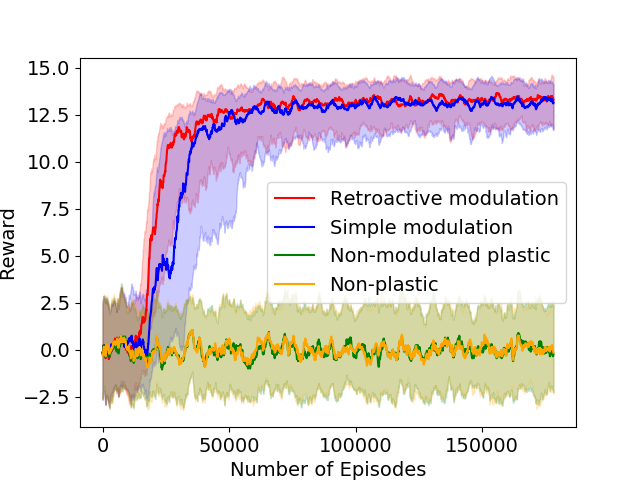}
\caption{Left: Description of the task. Right: Training curves for the cue-reward association task (medians and inter-quartile ranges of rewards per episode over 10 runs). Modulated plastic networks (red, blue) learn the task, while non-modulated and non-plastic networks (green, orange) fail. 
%\ks{More caption text should help describe the figure and explain what it tells us.  Also, font in graphic is a bit too small (probably easily solved by just scaling the figure), and graphic should be centered if it's in this format.}
}
\end{center}
\label{fig:sr}
\end{figure}

Our first test task is a simple meta-learning problem that emulates an animal behavioral learning task, as described in Figure \ref{fig:sr} (Left). In each episode, one of four input cues is arbitrarily chosen as the \emph{Target} cue. Repeatedly, the agent is shown two cues in succession, randomly chosen from the possible four, then a \emph{Response} cue during which the agent must respond 1 if the Target cue was part of the pair, or 0 otherwise. A correct response produces a reward of 1.0, while an incorrect response returns reward -1.0 (this is a two-alternative forced choice task: a response of either 1 or 0 is always produced). This process iterates for the duration of the episode, which is 200 time steps. The cues are binary vectors of 20 bits, randomly generated at the beginning of each episode. To prevent simple time-locked scheduling strategies, a variable number of zero-input time steps are randomly inserted, including at least one after each presentation of the Go cue; as a result, the length of each trial varies, and the number of trials per episode is somewhat variable (the mean number of trials per episode is 15).
 
%Note that the episode can thus be seen as being composed of a succession of \emph{trials}, where each trial consists of the presentation of a pair of cues, the Go cue, and delivery or not of reward (trials are only a division of convenience for describing the experiment; the agent only perceives a succession of time steps, with no specific input to mark trials beyond the Go cue). A variable number of zero-input time steps are randomly inserted in each trial (including at least one at the beginning of each trial, i.e. after each presentation of the Go cue); as a result, the length of each trial varies, and the number of trials per episode is somewhat variable (the mean number of trials per episode is 15). This was done to prevent simple time-locked scheduling strategies.

The architecture is a simple recurrent network with 200 neurons in the hidden recurrent layer. Only the recurrent layer is plastic: input and output weights are non-plastic, having only $w_{i,j}$ coefficients. There are 24 inputs: 20 binary inputs for the current cue and one input providing the time elapsed since the start of the episode, as well as two binary inputs for the one-hot encoded response at the previous time step and one real-valued channel for the reward received at the previous time step, in accordance with common meta-learning practice \citep{wang2016learning,duan2016rl2}. There are four outputs: two binary outputs for the one-hot encoded response, plus an output neuron that predicts the sum of future discounted rewards $V(t)$ over the remainder of the episode (as mandated by the A2C algorithm that we use for meta-training, following \citet{wang2016learning}), and the neuromodulatory signal $M(t)$. The two response outputs undergo a softmax operation to produce probabilities over the response, while the $M(t)$ signal is passed through a $\tanh$ nonlinearity and the $V(t)$ output is a pure linear output. All gradients are clipped at norm 7.0, which greatly improved stability.

Training curves are shown in Figure \ref{fig:sr} (Right; each curve shows the median and inter-quartile range over 10 runs).
%\ks{why 9 runs?  it would seem less strange if it were 10 runs} with different random seeds). 
Neuromodulatory approaches 
% I wanted an odd number because the median and IQR makes more sense then, but I'll switch back to 10
%\ks{significantly?} 
succeed in learning the task, while non-neuromodulatory networks \citep{miconi2016backpropagation,miconi2018differentiable}) and non-plastic, simple recurrent networks fail to learn it. \jc{I think we should cut this. It's dangerous for reviewers, as they think they are missing out on lots of the experimentation you did. I also dont' think it is needed. This is our theory, so we can just say it. The theory is based on these data and the previous paper too, which is clear without this caveat: Based on previous experimentation with different forms of this task, w}\tm{OK}We hypothesize that this dramatic difference is related to the relatively high dimensionality of the input cues: just as non-modulated plastic networks seemed to outperform non-plastic networks specifically when required to memorize arbitrary high-dimensional stimuli \citep{miconi2018differentiable}, neuromodulation seems to specifically help memorizing reward associations with such arbitrary high-dimensional stimuli (see Appendix). \ks{any hypothesis on why? (Jeff asks same in written comments)}\tm{Something extremely hand-wavy, based on comparing this result with the one of the previous version that used fixed, 4-bit cues (and in which non-modulated plasticity did much better). I'll move most of it to the Appendix.}\jc{Much better! (with the change I propose above of removing the first clause)}

To illustrate the behavior of neuromodulation, we plotted the output of the neuromodulator neuron for several trials. These graphs reveal that neuromodulation reacts to reward in a complex, time-dependent manner (see Appendix).

%\ks{Should we do a statistical test to support a more formal statement about significance?  "outperform" and "appear small" sounds a bit informal (I don't care myself but a reviewer might pick on us)}

%\ks{Nice that the NM versions perform better, but can you say anything (briefly) about why?  What explains this disparity in the particular case of this experiment?  More broadly, ideally the paper would at some point advance a clear hypothesis about *why* NM would improve performance in certain types of tasks (as opposed to with raw plasticity without it), and then here we could conclude by explaining how this result confirms that hypothesis.  The paper impact is stronger if the reader gets the *why* and not just the what.}

\subsection{Task 2: Maze navigation task}

\begin{figure}
\includegraphics[scale=.18]{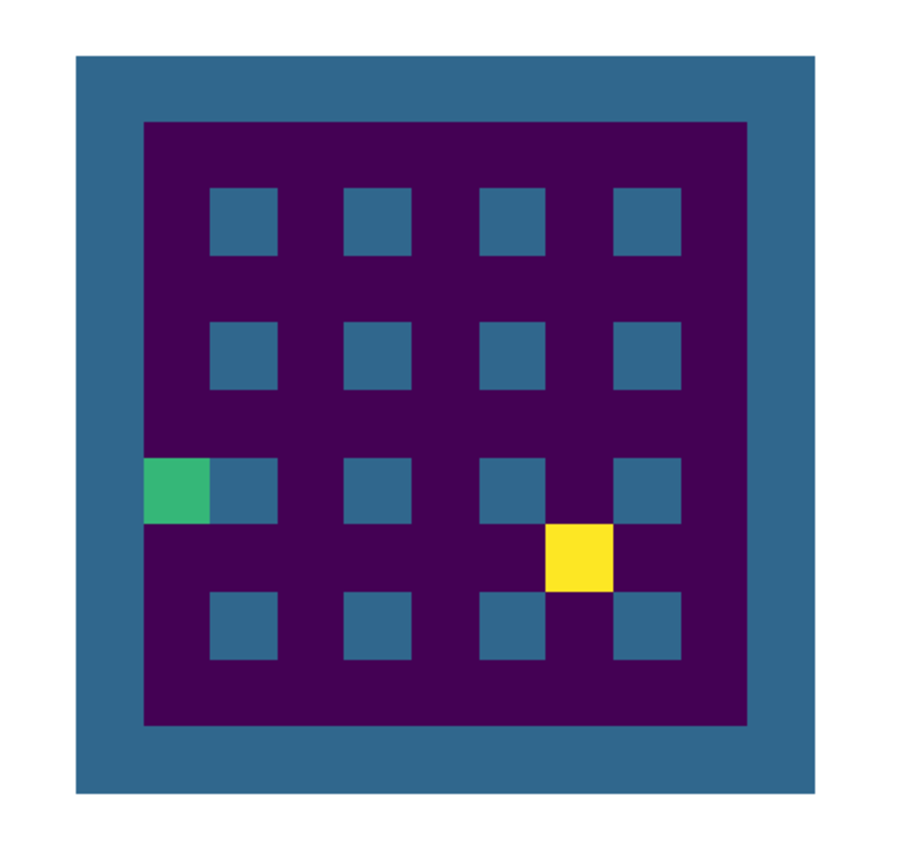}
\includegraphics[scale=.5]{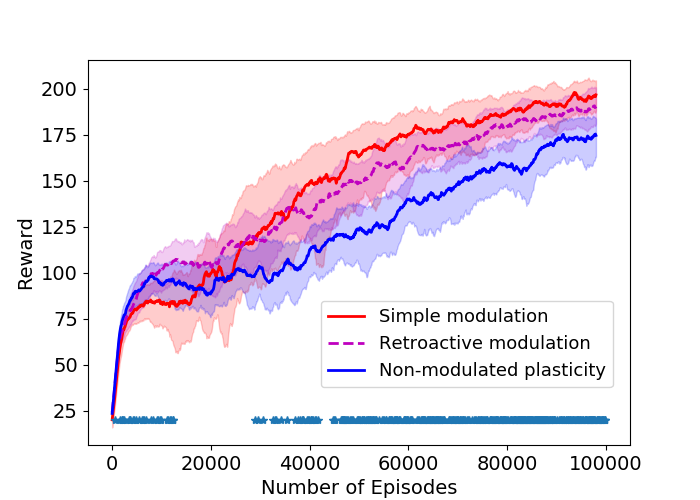}
\caption{Maze navigation task. Left: layout of the maze, including an example agent location (yellow) and reward location (green, for illustration only: the reward is not visible to the agent). Right: Training curves for the maze exploration task: median and inter-quartile range of reward over 9 runs for each episode. Cyan stars (bottom) indicate statistically significant difference between simple neuromodulation and non-modulated plasticity at $p<0.05$ (Wilcoxon rank-sum test).\todo{resize figures so they don't flow out into the margin. I have seen reviewers reject papers over such minutiae!}}\tm{Done}
\label{fig:maze}
\end{figure}

% The other task is also an RL task!
For a more challenging problem, we also tested the approach on 
the grid maze exploration task introduced by \citet{miconi2018differentiable}. Here, the maze is composed of $9 \times 9$ squares, surrounded by walls, in which every other square (in either direction) is occupied by a wall.
%(except for the central square, which is left empty because it is the starting agent location for each episode). 
Thus the maze contains 16 wall squares, arranged in a regular grid except for the center square (Figure \ref{fig:maze}, left). The shape of the maze is fixed and unchanging over the whole task. At each episode, one non-wall square is randomly chosen as the reward location. When the agent hits this location, it receives a reward and is immediately transported to a \emph{random} location in the maze. Each episode lasts 200 time steps, during which the agent must accumulate as much reward as possible. The reward location is fixed within an episode and randomized across episodes. Note that the reward is invisible to the agent, and thus the agent only knows it has hit the reward location by the activation of the reward input at the next step (and possibly by the teleportation, if it can detect it).

The architecture is the same as for the previous task, but with only 100 recurrent neurons. The outputs consist of 4 action channels (i.e. one for each of the possible actions: left, right, up or down) passed through a softmax, as well as the pure linear $V(t)$ output and the $M(t)$ neuromodulatory signal passed through a $\tanh$ nonlinearity.  Inputs to the agent consist of a binary vector describing the $3 \times 3$ neighborhood centered on the agent (each element being set to 1 or 0 if the corresponding square is or is not a wall), plus four additional inputs for the one-hot encoded action taken at the previous time step, and one input for the reward received at the previous time step, following common practice \citep{wang2016learning}. Again, only recurrent weights are plastic: input-to-recurrent and recurrent-to-output weights are non-plastic. Results in Figure \ref{fig:maze} show that modulatory approaches again outperform non-modulated plasticity. 

%These results suggest that gradient descent successfully exploited the additional flexibility afforded by self-modulation of plasticity: the ability to choose not just which connections should be plastic, but when they should be plastic, improved the network's learning performance for these two reinforcement learning tasks.

%\ks{again, any chance we can talk about significance?}
%\ks{again, any insight into why they did better?  what's the hypothesis on why it would be or should be?}
% Any 

%\ks{as before, I notice we say nothing on significance.  is there a reason for that?  we think this community won't care?  it's evident anyway from inter-quartile range markings?  maybe there is a reason we don't need it, but i want to make we've thought about it.  we use terms like "small, and no obvious" regarding clipping, which raises the question what gives us the right to make judgements based on such subjective terms as being "obvious" - is that also the standard for concluding that NM works better than non-modulatory nets?}\jc{I think we should always say whether a result is statistically significantly different or not, including listing the statistical test used and the p value.}

\subsection{Task 3: Language Modeling}

Word-level language modeling is a supervised learning sequence problem, where the goal is to predict the next word in a large language corpus. Language modeling requires storing long term context, and therefore LSTM models \citep{Hochreiter97longshort-term} generally perform well on this task \citep{DBLP:journals/corr/ZarembaSV14}. The goal of this experiment is to study the benefits of adding plasticity and neuromodulation to LSTMs. 
%Recent studies have shown that adding so-called ``fast weights''

The Penn Tree Bank corpus (PTB), a well known benchmark for language modeling \citep{Marcus:1993:BLA:972470.972475}, is used here for comparing different models. The dataset consists of 929k training words, 73k validation words, and 82k, test words, with a vocabulary of 10k words. 

%During training successive mini-batches of size 20 are sequentially processed. 

For this task, we implemented neuromodulated plasticity in two different models: a basic model with 4.8 million parameters inspired from \citet{DBLP:journals/corr/ZarembaSV14}, and a much larger and more complex model with 24.2 million parameters, forked from \citet{merity2017regularizing}. The smaller model allowed for more experimentation, while the larger model showcases the results of neuromodulation on a complex model with (at the time of writing) state-of-the-art performance.

Detailed experimental descriptions are provided in the Appendix, summarized here: For the basic model, each network consists of an embedding layer, followed by two LSTM layers (approximately of size 200). The size of the LSTM layers is adjusted to ensure that the total number of trainable parameters remains constant across all experiments; note that this includes all plasticity-related additional parameters, i.e. $\alpha_{i,j}$ as well as the additional parameters related to neuromodulation (see Appendix). The final layer is a softmax layer of size 10k. The network is unrolled for 20 time steps during backpropagation through time ~\citep{werbos1990backpropagation}. The norm of the gradient is clipped at 5. This setup is similar to the non-regularized model  described by \citet{DBLP:journals/corr/ZarembaSV14}. One difference is that an extra L2 penalty is added to the weights of the network here (adding this penalty consistently improves results for all the models).  

The large model, as described in \citep{merity2017regularizing}, consists of an embedding of size 400 followed by three LSTMs with 1150, 1150 and 400 cells respectively (which we reduce to 1149, 1149 and 400 for the plastic version only to ensure the total number of trainable parameters is not higher). Importantly \citet{merity2017regularizing} use various regularization techniques. For example, the training process involves switching from standard SGD to Averaged-SGD  after a number of epochs. The main departures from \citet{merity2017regularizing} is that we do not implement recurrent dropout (feedforward dropout is preserved) and reduce batch size to 7 due to computational limitations. Other hyperparameters are taken ``as is'' without any tuning. See Appendix for details.

\begin{table}[t]
\caption{Test Perplexity Results on Penn-Tree Bank. Lower values are better. Basic model: mean and 95\% CI over 16 runs. Large model: median and (min, max) over 5 runs.
%For the smaller model, retroactive neuromodulation outperforms baseline LSTMs by 1.7 perplexity point ($p<1e-7$, Wilcoxon rank-sum test)
}
\label{table:language}
\medskip
%\ks{note that if we end up needing to cut, this caption is a possible target: the text about differences between treatments here is almost entirely redundant with the text at the end of the section. Also, we could put our top result in boldface within the table if we want.}\jc{agree with Ken on both points} % title of Table

\centering % used for centering table
\begin{tabular}{c c} % centered columns (2 columns)
\hline\hline %inserts double horizontal lines
\textbf{Model} & \textbf{Test Perplexity} \\ [0.5ex] % inserts table
%heading
\hline % inserts single horizontal line
Baseline LSTM (similar to \citet{DBLP:journals/corr/ZarembaSV14}) & 104.26 $\pm$ 0.22 \\ % inserting body of the table
LSTM with Differential Plasticity & 103.80 $\pm$ 0.25 \\
LSTM with Simple Neuromodulation & 102.65 $\pm$ 0.30\\
LSTM with Retroactive Neuromodulation & 102.48 $\pm$ 0.28 \\
\hline
Baseline large LSTM model (from \citet{merity2017regularizing}) & 62.48 (62.40, 62.60) \\
Large LSTM model with neuromodulated plasticity & 61.44 (61.37, 61.68)\\
[1ex] % [1ex] adds vertical space
\hline %inserts single line
\end{tabular}
%\caption{Test Perplexity Results on Penn-Tree Bank. Shown are the mean perplexity values (over 16 runs) with the 95\% confidence interval. Lower perplexity values are better. Each model has the same number of total parameters (4.8 Million). Adding differentiable plasticity to LSTMs provides small, but statistically significant gains over Baseline LSTMs ($p=0.0044$). Making the plastic connections neuromodulated yields a $\sim$1.7 perplexity point improvement over Baseline LSTMs, which is significantly higher-performing than both non-neuromodulated, plastic LSTMs ($p=1e-6$) and Baseline LSTMs ($p=1e-7$). The difference between simple and retroactive neuromodulation is not statistically significant ($p=0.066$).
%\ks{note that if we end up needing to cut, this caption is a possible target: the text about differences between treatments here is almost entirely redundant with the text at the end of the section. Also, we could put our top result in boldface within the table if we want.}\jc{agree with Ken on both points} % title of Table
%}
\end{table}

Four versions of the smaller model are evaluated here (Table~\ref{table:language}). (1) The \emph{Baseline LSTM} model (described in the previous paragraph)\footnote{The Baseline LSTM performance is better than the one published in \citep{DBLP:journals/corr/ZarembaSV14} due to our hyperparameter tuning, as described in the Appendix.}. (2) \emph{LSTM with Differentiable Plasticity}: there are four recurrent connections in each LSTM node and here, plasticity is added to one of them (see ~\ref{appendix:plasiclstm} for details) as per equations ~\ref{eq:network} and ~\ref{eq:hebb}. Because the number of plastic connections is large, each plastic connection has its own individual $\eta$ so that their values can be individually tuned by backpropagation. (3) \emph{LSTM with Simple Neuromodulation:} here simple neuromodulation is introduced following equation \ref{eq:simplemodulation}. The $\eta$ parameters are replaced by the output of a neuron $M(t)$. $M(t)$ itself receives as input a weighted combination of the hidden layer's activations, where the weights are learned in the usual way. There is one $M(t)$ associated with each LSTM layer. (4) \emph{LSTM with Retroactive Neuromodulation}: this model is the same as the LSTM with Simple Neuromodulation, except it uses the equations that enable eligibility traces (equations \ref{eq:retroactive1} and \ref{eq:retroactive2}). Additional details for the plastic and neuromodulated plastic LSTMs are described in the Appendix.  

For each of the four models, we separately searched for the best hyperparameters with equally-powered grid-search. Each model was then run 16 times with its best hyperparameter settings. The mean test perplexity of these 16 runs along with the 95\% confidence interval is presented in Table ~\ref{table:language}. Results show that adding differentiable plasticity to LSTM provides slightly, but statistically significantly better results than the Baseline LSTM (Wilcoxon rank-sum test, $p=0.0044$). Adding neuromodulation further (and statistically significantly) lowers the perplexity over and above the LSTM with differential plasticity ($p=1e-6$). Overall, retroactive neuromodulation provides about 1.7 perplexity improvement vs. the Baseline LSTM (statistically significant, $p=1e-7$). Retroactive neuromodulation (i.e.\ with eligibility traces) does outperform simple neuromodulation, but the improvement is just barely not statistically significant at the traditional $p<0.05$ cutoff ($p=0.066$). Note that while these figures are far from state-of-the-art results (which use considerably larger, more complex architectures), they all still outperform published work using similar architectures \citep{DBLP:journals/corr/ZarembaSV14}.

For the larger model, we compare a version in which the core LSTM module has been reimplemented to have neuromodulated plastic connections (simple neuromdulation only; no retroactive modulation was implemented), and a baseline model that uses the same LSTM reimplementation but without the plasticity and modulation, in order to make the comparison as equal as possible. Note that in this model, plasticity coefficients are attributed ``per neuron'': there is only one $\alpha_i$ for each neuron $i$ (as opposed to one per connection), which is applied to all the Hebbian traces of the connections incoming to this neuron. This helps limit the total number of parameter. See the Appendix for a more complete description. The modulated plastic model shows a small improvement over the non-plastic version (Table \ref{table:language}), confirming the results obtained with the smaller model.

%KEN: edited a little
%These results indicate that LSTMs can perform better when combined with plasticity and (even more so) neuromodulated plasticity. They also demonstrate the benefits of these two forms of plasticity in a supervised learning setting, specifically a standard benchmark language modeling task. 

\section{Discussion and Future Work}

This paper introduces a biologically-inspired method for training networks to self-modify their weights\jc{their weights}. Building upon the differentiable plasticity framework, which already improved performance (sometimes dramatically) over non-plastic architectures on various supervised and RL tasks \citep{miconi2016backpropagation,miconi2018differentiable}, 
%which made (passive) plasticity amenable to gradient descent, which already demonstrated improved meta-learning performance over non-plastic networks,
\ks{again we may want to remind the reader that passive plasticity yielded some kind of demonstrable advatange -- otherwise, why should we care if neuromodulation improves upon it?  what does it matter that you can get neuromodulated plasticity if plasticity itself has no empirical benefit in neural networks?}\jc{agreed. we have to sell them on plasticity being interesting and useful first, then we can say NM is even more powerful. We can use our motivation written for the last paper to do that (reworded)}\tm{I've added a short mention above. Also, note that we are including results with non-plastic networks in the present paper (except for the maze task because I don't have the data yet).}\jc{I like the new version.}\ks{like it too, I think it resolves the issue.} here we introduce neuromodulated plasticity to let the network control its own weight changes. As a result, for the first time, neuromodulated plastic networks can be trained with gradient descent, opening up a new research direction into optimizing large-scale self-modifying neural networks. 

%This framework thus combines one of the most powerful tools in the toolbox of biological learning in brains -- neuromodulated plasticity -- with one of the most powerful tools in the toolbox of machine learning -- gradient descent. It thus opens up a new research direction into optimizing large-scale self-modifying neural networks.

As a complement to the benefits in the simple RL domains investigated,
our finding that plastic and neuromodulated LSTMs outperform standard LSTMs on a benchmark language modeling task (importantly, a central domain of application of LSTMs) is potentially of great importance. LSTMs are used in real-world applications with massive academic and economic impact. Therefore, if plasticity and neuromodulation consistently improve LSTM performance (for a fixed search space size), the potential benefits could be considerable. We intend to pursue this line of investigation and test plastic LSTMs (both neuromodulated and non) on other problems for which LSTMs are commonly used, such as forecasting.\jc{this would be more powerful if we listed a few more, and more mainstream, killer app domains of LSTMs. Translation? Speech-to-text? Others?}

Conceptually, an important comparison point is the ``Learning to Reinforcement Learn'' (L2RL) framework introduced by \citet{wang2016learning,wang2018prefrontal}. In this meta-learning framework, the weights do not change during episodes: all within-episode learning occurs through updates to the activity state of the network. This framework is explicitly described \citep{wang2018prefrontal} as a model of the slow sculpting of prefrontal cortex by the reward-based dopamine system, an analogy facilitated by the features of the A2C algorithm used for meta-training  (such as the use of a value signal and modulation of weight changes by a reward prediction error). As described in the RL experiments above, our approach adds more flexibility to this model by allowing the system to store state information with weight changes, in addition to hidden state changes. However, because our framework allows the network to update its own connectivity, we might potentially extend the L2RL model one level higher: rather than using A2C as a hand-designed reward-based weight-modification scheme, the system could now determine its own arbitrary weight-modification scheme, which might make use of any signal it can compute (reward predictions, surprise, saliency, etc.) This emergent weight-modifying algorithm (designed over many episodes/lifetimes by the ``outer loop'' meta-training algorithm) might in turn sculpt network connectivity to implement the meta-learning process described by \citet{wang2018prefrontal}. 
Importantly, this additional level of learning (or ``meta-meta-learning'') is not just a pure flight of fancy: it has undoubtedly taken place in evolution.
Because  humans (and other animals) can perform meta-learning (``learning-to-learn'') during their lifetime \citep{harlow1949formation,wang2018prefrontal}, and because humans are themselves the result of an optimization process (evolution), then meta-meta-learning has not only occurred, but may be the key to some of the most advanced human mental functions. Our framework opens the tantalizing possibility of studying this process, while allowing us to replace evolution with any gradient-based method in the outermost optimization loop.
\jc{I think optimization would work better than design here, although I know Ken does not like that word. I think we'll confuse people a bit with the current phrasing.}\ks{the wording here is from Thomas.  I only object to "optimization" in the context of open-ended systems.  anyway, I don't have strong feelings here.}
\jc{given that, I suggest we change to the more conventional (and thus clearer) ``outermost optimization loop''}\tm{Done}

\ks{The final paragraph focuses heavily on next-level RL meta-meta stuff, but (echoing my previous comment) we may want to branch somewhere into the implications of improved language modeling and time series predicton through neuromodulation, which may require a different narrative.  Ultimately, weaving the two narratives together seamlessly is a challenge for wrapping up the paper satisfyingly and highlighting the marquee result of beating a raw LSTM!}
\tm{I have tried to generalize it a bit and added a paragraph for Aditya's result above}\jc{I like the new paragraph Thomas added. I think ``More conceptually'' (the transition after it) is vague, but if we can't think of anything better it's fine.}
\ks{To address these issues, I changed some of the transitional text on both paragraphs to make it a bit less jarring.  First, jumping straight into a discussion of the LSTM result *first* when its experiment was *last* felt incongruous, so I put a little text in front to acknowledge the other results first.  Second, I changed "More conceptually" to "At a conceptual level," which I think makes it more clear why we are making the transition, that is, we are moving from a discussion of empirical implications to a conceptual discussion, as opposed to just getting "more conceptual," which is vague.}\jc{I like it better now}\tm{OK}

To investigate the full potential of our approach, the framework described above requires several improvements. These include: implementing multiple neuromodulatory signals (each with their own inputs and outputs\jc{what exactly does this mean? literally different inputs? or different weights on the same inputs? how about, ``each with a different learned function of the inputs (i.e. triggered by different events)}\jc{Aditya's model already has two $M(t)$s...so doesn't the paper already have this now?}), as seems to be the case in the brain \citep{lammel2014reward,howe2016rapid,saunders2018dopamine}; introducing more complex tasks that could make full use of the flexibility of the framework, including the eligibility traces afforded by retroactive modulation and the several levels of learning mentioned above\jc{before we get to using the many levels, we could mention testing these new models on many more problems to understand their pros and cons, and also understand when the eligibility trace version has a substantial advantage over the simple neuromodulation version, as we did not see a difference in any of the problems we studied, despite their being good theoretical and empirical reasons to expect it should perform better in many cases}\ks{yes an acknowledgement of the potential for the eligibility trace section would leave it seeming less lonely among the more proven parts of the paper}\tm{Tried to do that!}; and addressing the pitfalls in the implementation of reinforcement learning with reward-modulated Hebbian plasticity (e.g.\ the inherent interference between the unsupervised component of Hebbian learning and reward-based modifications; \citealt{fremaux2010functional,fremaux2015neuromodulated}), so as to facilitate the automatic design of efficient, self-contained reinforcement learning systems\jc{I don't think most readers will understand this, though at least they know they can go read more if they want. Given the late hour, this is fine for ICLR...but if there is a quick way to explain this more clearly, it woudl strengthen the paper}. Finally, it might be necessary to allow the meta-training algorithm to design the overall architecture of the system, rather than simply the parameters of a fixed, hand-designed architecture. With such a rich potential for extension, our framework for neuromodulated plastic networks opens many avenues of exciting research.

\bibliography{biblio}
\bibliographystyle{iclr2019_conference}

\appendix

\section{Appendix}

\subsection{Plastic LSTMs: basic model}
\label{appendix:plasiclstm}

\subsubsection{Adding plasticity to LSTMs}

Each LSTM node consists of four weighted recurrent paths through $i_t$, $j_t$, $f_t$ and $o_t$ as shown in the equations below: 

\begin{equation}\label{eq:i_t}
        i_t = \tanh(W_{xi}x_t + W_{hi}h_{t-1} + b_i)
\end{equation} 
\begin{equation}
        j_t = \sigma(W_{xj}x_t + W_{hj}h_{t-1} + b_j)\\
\end{equation}
\begin{equation}
        f_t = \sigma(W_{xf}x_t + W_{hf}h_{t-1} + b_f)\\
\end{equation}
\begin{equation}
        o_t = \sigma(W_{xo}x_t + W_{ho}h_{t-1} + b_o)\\
\end{equation}
\begin{equation}
        c_t = f_t \otimes c_{t-1} + i_t \otimes j_t\\
\end{equation}
\begin{equation}
        h_t = \tanh(c_t) \otimes o_t
\end{equation}

 $j_t$, $f_t$ and $o_t$ are used for controlling the data-flow through the LSTM and $i_t$ is the actual data. Therefore, plasticity is introduced in the path that goes through $i_t$ (adding plasticity to the control paths of LSTM is for future-work) . The corresponding pre-synaptic and post-synaptic activations (denoted by $x_i(t-1)$ and $x_j(t)$ respectively in equations ~\ref{eq:network} and ~\ref{eq:hebb}) are $h_{t-1}$ and $i_t$. A layer of size 200 has 40k (200$\times$200) plastic connections. Each plastic connection has its own individual $\eta$ (used in equation ~\ref{eq:hebb}) that is learned through backpropagation. The plasticity coefficients ($\alpha_{i,j}$) are used as shown in equation~\ref{eq:network}. 

%Interestingly, unlike the fast-weights LSTM (~\citep{DBLP:journals/corr/abs-1804-06511}, there was no need to include any extra normalization layers to stabilize the LSTM activations. 

\subsubsection{Adding neuromodulation to LSTMs}

As shown in equation~\ref{eq:simplemodulation}, for simple neuromodulation, the $\eta$ is replaced by the output of a network computed neuron $M(t)$. For neuromodulated LSTMs, individual $\eta$ for each plastic connection is replaced by the output of a neuron ($M(t)$) that has a fan-out equal to the number of plastic connections. The input to this neuron is the activations $h_{t-1}$ of the layer from the previous time-step. Each LSTM layer has its dedicated neuromodulatory neuron. Other variations of this setting include having one dedicated neuromodulatory neuron per node or having one neuromodulatory neuron for the whole network. Preliminary experiments showed that these variations performed worse and therefore they were not further evaluated.

\subsection{More details for Language Modeling experiment}
All the four models presented in Table ~\ref{table:language} are trained using SGD. Initial learning rate was set 1.0. Each model is trained for 13 epochs. The hidden states of LSTM are initialized to zero; the final hidden states of the current minibatch are used as the initial hidden states of the subsequent minibatch. 

Grid-search was performed for four hyperparameters: (1) Learning rate decay factor in the range $0.25$ to $0.4$ in steps of $0.01$. (2) Epoch at which learning rate decay begins in the range - \{$4, 5, 6$\}. (3) Initial scale of weights in the range - \{$0.09, 0.1, 0.11, 0.12$\}. (4) L2 penalty constant in the range - \{$1e-2, 1e-3, 1e-4, 1e-5, 1e-6$\}. 

%The state-of-art results in language modeling domain have been achieved with much larger LSTM models \cite{melis2018on}. In such large models, recurrent dropouts \citep{rnndropout} need to be introduced for regularization. An interesting direction for future-work is to understand the effects of dropout on plastic and neuromodulated connections. 

\subsection{Large word-modelling network}

In addition to the previous model, we also applied the Backpropamine framework to the much larger, state-of-the-art model described by \citet{merity2017regularizing}. This model consists of three stacked LSTMs with 115, 1150 and 400 cells each, with an input embedding of size 400 and an output softmax layer that shares weights with the input embedding. The model makes use of numerous optimization and regularization techniques. Connections between successive LSTMs implement ``variational'' dropout, in which a common dropout mask is used for the entire forward and backward pass \cite{gal2016theoretically}. Backpropagation through time uses a variable horizon centered on 70 words. After 45 epochs, the optimizer switches from SGD (without momentum) to Averaged-SGD, which consists in computing standard SGD steps but taking the average of the resulting successive updated weight vectors. This is all in accordance with \citet{merity2017regularizing}. The only differences are that we do not implement weight-dropout in recurrent connections, force the switch to ASGD at 45 epochs for all runs of all models, and limit batch size to 7 due to computational restrictions. 

Plasticity coefficients are attributed ``per neuron'': rather than having and independent $\alpha_{i,j}$ for each connection, each neuron $i$ has a plasticity coefficient $\alpha_i$ that is applied to all its incoming connection (note that Hebbian traces $\mathrm{Hebb}_{i,j}$ are still individually maintained for each connection). This reduces the number of trainable parameters, since $\alpha$  is now a vector of length $N$ rather than a matrix of size $N \times N$ (where $N$ is the number of recurrent neurons). 

We implement simple neuromodulation as described in Equation \ref{eq:simplemodulation}. A single neuromodulator neuron with $\tanh$ nonlinearity receives input from all recurrent neurons. This neuromodulator input is then passed through a vector of weights, one per neuron, to produce a different $\eta_i$ for each neuron. In other words, different neurons $i$ have different $\eta_i$, but these are all fixed multiples of a common value. This is an intermediate solution between having a single $\eta(t)$ for the whole network, and independently computing a separate $\eta_i$ for each neuron, each with its own input weights (which would require $N \times N$ weights, rather than $2 \times N$ for the current solution). Neuromodulation is computed separately to each of the three LSTMs in the model.

For the non-plastic network, the total number of trainable parameters is 24\,221\,600. For the neuromodulated plastic version, we reduce the number of hidden cells in LSTMs from 1150 to 1149, which suffices to bring the total number of parameters down to 24\,198\,893 trainable parameters (rather than 24\,229\,703 for 1150-cell LSTMs).

All other hyperparameters are taken from \citet{merity2017regularizing}, using the instructions provided on the code repository for their model, available at \url{https://github.com/salesforce/awd-lstm-lm}. We did not perform any hyperparameter tuning due to computational constraints.

\subsection{Dynamics of neuromodulation}

\begin{figure}
\includegraphics[scale=.4]{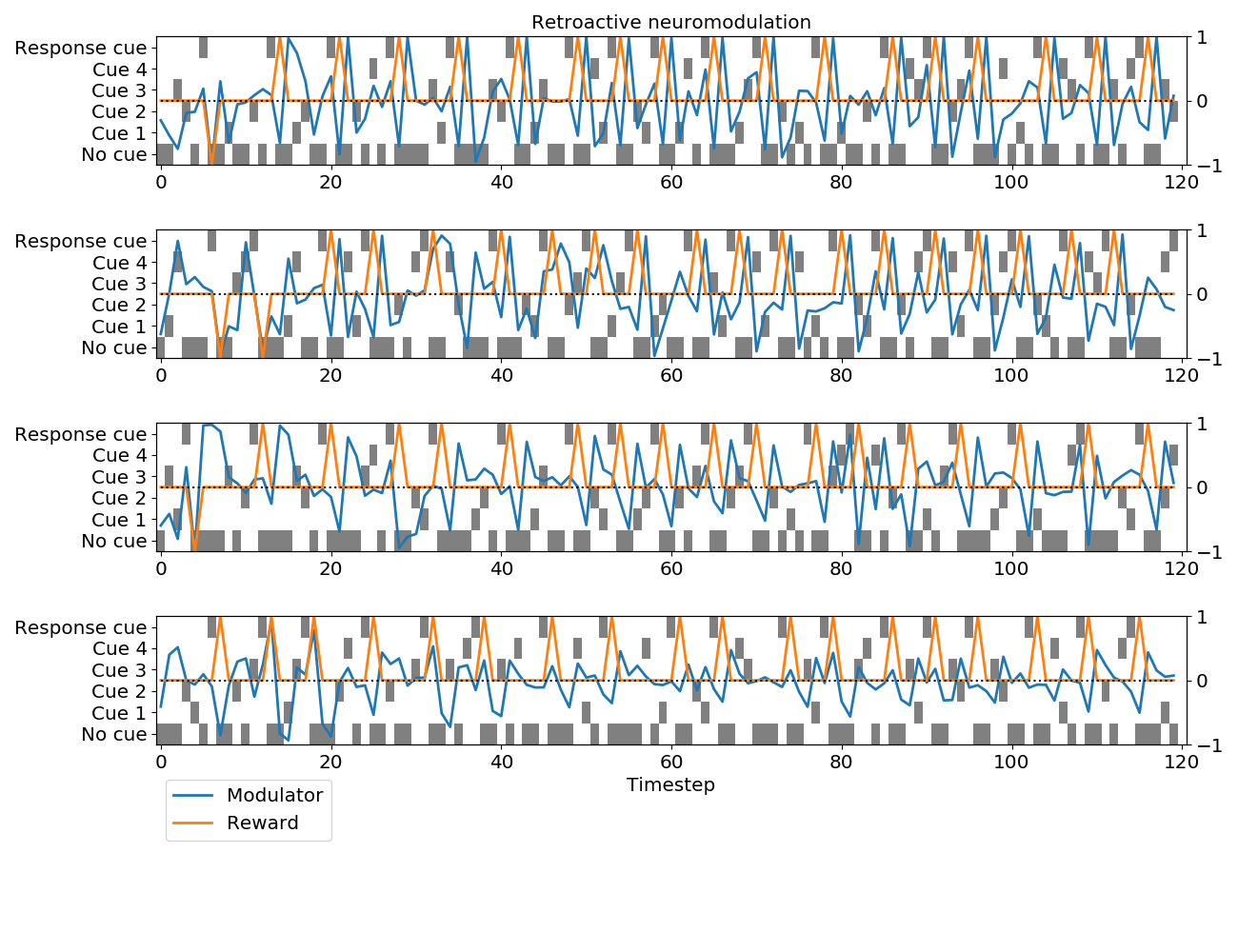}
\includegraphics[scale=.4]{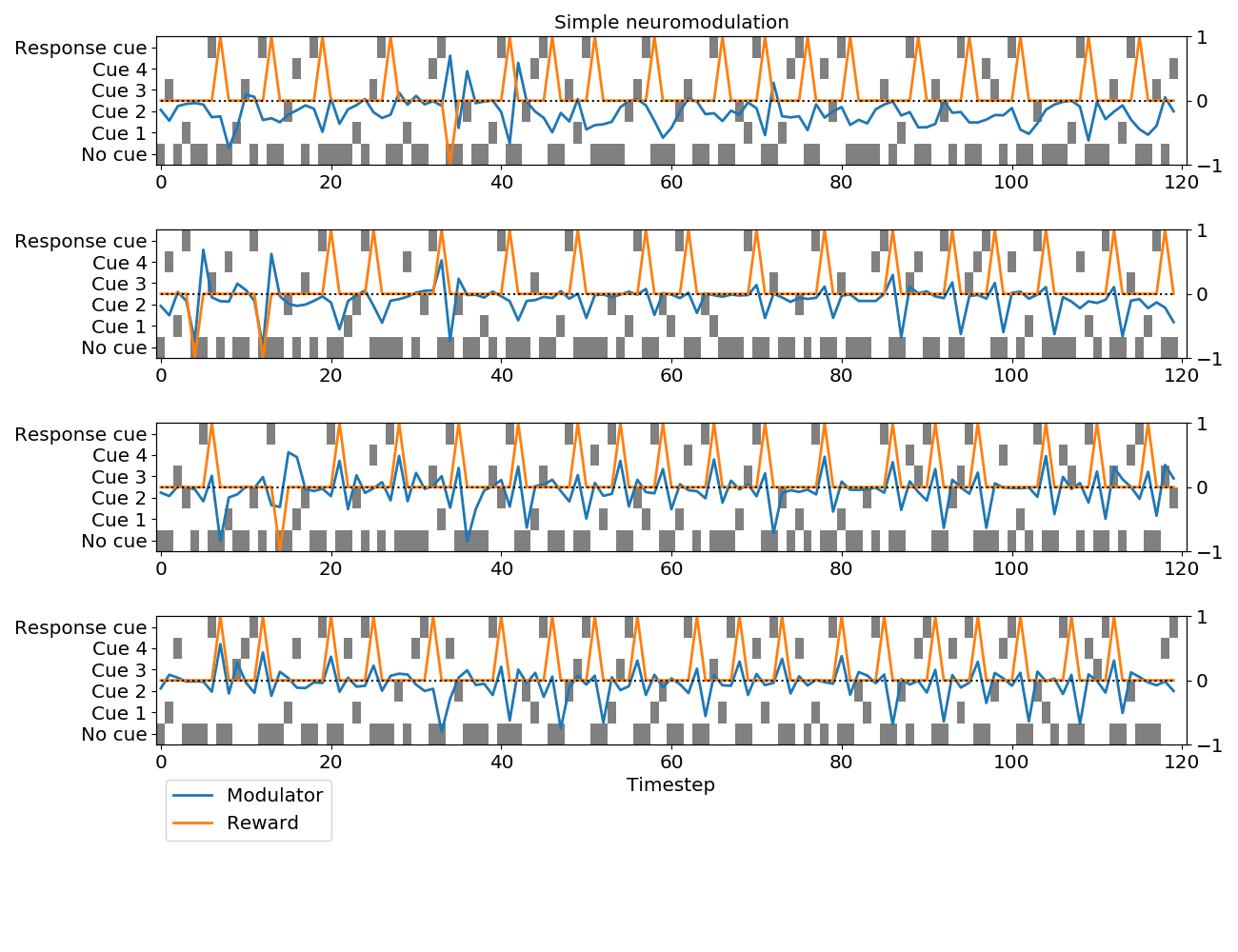}
\caption{Dynamics of neuromodulation. For both simple and retroactive modulation, we show one trial from each of 4 runs. At each time step, gray squares indicate which cue was shown; orange curves indicates reward at the previous time step (i.e. the one currently fed to the network), which is always -1, 0 or 1; and blue curves indicates the value of the modulator output at the current time step. Notice the diversity of dynamics.}
\label{fig:modulation}
\end{figure}

To illustrate the behavior of neuromodulation, we plot the output of the neuromodulator neuron for random trials from several runs of Task 1 (Figure \ref{fig:modulation}). All runs are from well-trained, highly successful network, as seen by the low proportion of negative rewards. For each run, we plot both the value of the neuromodulator output at each time step, and the reward being currently perceived by the network (i.e. the one produced by the response at the previous time step). 

The plots reveal rich, complex dynamics that vary greatly between runs. The modulator neuron clearly reacts to reward; however, this reaction is complex, time-dependent and varies from run to run. The topmost run for retroactive modulation tends to produce negative neuromodulation in response to positive reward, and vice-versa; while the second-to-last run for simple neuromodulation tends to to the opposite. A common pattern is to produce negative neuromdulation on the time step just following reward perception (especially for simple neuromodulation). Two of the runs for retroactive modulation exhibit a pattern where reward perception is followed by highly positive, then highly negative neuromodulation. Understanding the mechanism by which these complex dynamics perform efficient within-episode learning is an important direction for future work.

\subsection{Cue-reward association task}

\begin{figure}
\begin{center}
\includegraphics[scale=.4]{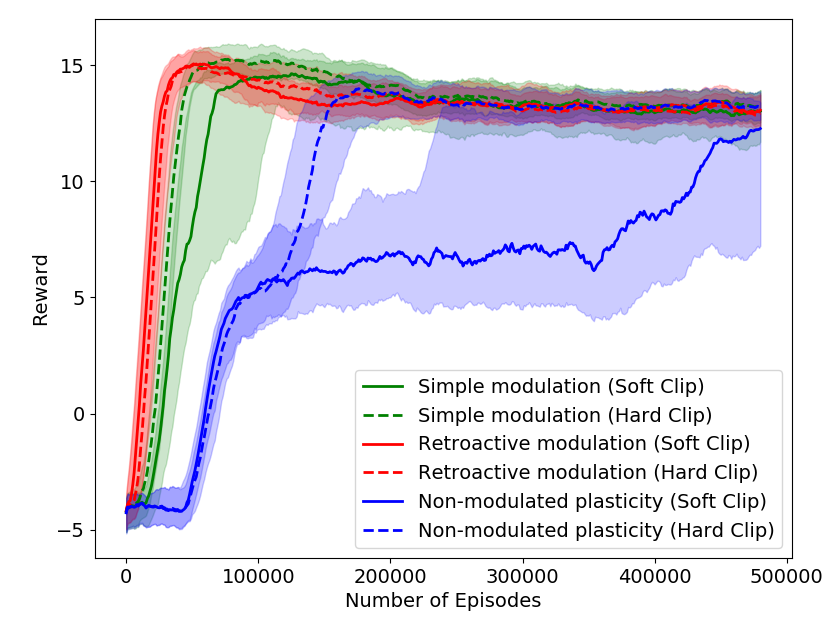}
\caption{Training curves for the cue-reward association task with fixed, binary four-bit cues (medians and inter-quartile ranges of rewards per episode over 10 runs). ``Soft clip'' refers to a different clipping operation used in Equation \ref{eq:hebb}; ``Hard clip'' is the same as used in the present paper, i.e. the simple clipping described in Methods.  Note that non-modulated plastic network succeed in solving this task. 
%\ks{More caption text should help describe the figure and explain what it tells us.  Also, font in graphic is a bit too small (probably easily solved by just scaling the figure), and graphic should be centered if it's in this format.}
}
\label{fig:sr_prev}
\end{center}
\end{figure}

In the cue-reward association learning task described above, neuromodulated plasticity was able to learn a task that non-modulated plasticity simply could not. What might be the source of this difference? In a previous experiment, we implemented the same task, but using only four fixed 4-bit binary cues for the entire task, namely, '1000', '0100', '0010' and '0001'. In this simplified version of the task, there is no need to memorize the cues for each episode, and the only thing to be learned for each episode is which of the four known cues is associated with reward. This is in contrast with the version used in the paper above, in which the cues are arbitrary 20-bits vectors randomly generated for each episode. With the fixed, four-bit cues, non-modulated plasticity was able to learn the task, though somewhat more slowly than neuromodulated plasticity (see Figure \ref{fig:sr_prev}).

This suggests neuromodulated plasticity could have a stronger advantage over non-modulated plasticity specifically in situations where the association to be learned involves arbitrary high-dimensional cues, which must be memorized jointly with the association itself. This echoes the results of \citet{miconi2018differentiable}, who suggest that plastic networks outperform non-plastic ones specifically on tasks requiring the fast memorization of high-dimensional inputs (e.g. image memorization and reconstruction task in \citep{miconi2018differentiable}).

Clearly, more work is needed to investigate which problems benefit most from neuromodulated plasticity, over non-modulated or non-plastic approaches. We intend to pursue this line of research in future work.

\end{document}